\pdfoutput=1

\documentclass[11pt]{article}

\usepackage[]{EMNLP2022}
\usepackage{array}
\usepackage{times}
\usepackage{latexsym}

\usepackage[T1]{fontenc}

\usepackage[utf8]{inputenc}

\usepackage{microtype}

\usepackage{inconsolata}

\usepackage{booktabs}
\usepackage{multirow}
\usepackage{rotating}
\usepackage{paralist}
\usepackage{comment} 
\usepackage[normalem]{ulem}
\usepackage{tipa}

\title{Leveraging Affirmative Interpretations from Negation\\Improves Natural Language Understanding}

\author{Md Mosharaf Hossain\mbox{\normalfont}\textsuperscript{\textipa{8,U}}\thanks{\ \ Work was done prior to joining Amazon.} \and        
        Eduardo Blanco\textsuperscript{\textipa{7}}\\
\textsuperscript{\textipa{8}}Department of Computer Science and Engineering, University of North Texas\\
\textsuperscript{\textipa{U}}Amazon\\
\textsuperscript{\textipa{7}}Department of Computer Science, 	University of Arizona\\
{\footnotesize
\texttt{mdmosharafhossain@my.unt.edu} \hspace{.2cm}
\texttt{hosmdmos@amazon.com} \hspace{.2cm}
\texttt{eduardoblanco@arizona.edu}}
}  

\begin{document}
\maketitle
\begin{abstract}
Negation poses a challenge in many natural language understanding tasks.
Inspired by the fact that understanding a negated statement often requires humans to infer affirmative interpretations,
in this paper we show that doing so benefits models for three natural language understanding tasks.
We present an automated procedure to collect pairs of sentences with negation and their affirmative interpretations,
resulting in over 150,000 pairs.
Experimental results show that leveraging these pairs helps
(a)~T5 generate affirmative interpretations from negations in a previous benchmark, and
(b)~a RoBERTa-based classifier solve the task of natural language inference.
We also leverage our pairs to build a plug-and-play neural generator that given a negated statement generates an affirmative interpretation.
Then, we incorporate the pretrained generator into a RoBERTa-based classifier for sentiment analysis and show that doing so improves the results.
Crucially, our proposal does not require any manual effort.


\end{abstract}

\section{Introduction}
\label{s:introduction}

Natural Language Understanding is a crucial component to build intelligent systems that interact with humans seamlessly.
While recent papers sometimes report so-called superhuman performance,
simple adversarial attacks including adding negation and other input modifications remain a challenge despite they are obvious to humans~\cite{naik-etal-2018-stress,wallace-etal-2019-universal}. 
Further, many researchers have found that state-of-the-art systems struggle with texts containing negation.
For example, \newcite{kassner-schutze-2020-negated} show that pretrained language models such as BERT~\cite{devlin-etal-2019-bert}
do not differentiate between negated and non-negated cloze questions (e.g., \emph{Birds cannot [MASK]} vs. \emph{Birds can [MASK]}). 
Other studies show that transformers perform much worse in many other natural language understanding tasks when there is a negation in the input sentence~\cite{
ribeiro-etal-2020-beyond,
ettinger-2020-bert,
hossain-etal-2020-analysis,
hosseini-etal-2021-understanding, 
hossain-etal-2022-analysis,
truong2022not}. 


\begin{figure}
\small
\centering
\begin{tabular}{@{}p{\columnwidth}@{}}
\toprule
English-Norwegian (en-no) parallel sentences: \\	
(en) \emph{There is \underline{no} more than one Truth.} \\
(no) \emph{Og det finnes kun en Sannhet.} \\
Backtranslation: \emph{And there is only one truth.} \\ \midrule
    
English-Spanish (en-es) parallel sentences: \\
(en) \emph{The term gained traction only after 1999.} \\
(es) \emph{El término \underline{no} se popularizó hasta después del 1999.} \\
Backtranslation: \emph{The term was \underline{not} popular until 1999.} \\ \bottomrule 
\end{tabular}
\caption{Parallel sentences from bitext corpora (English-Norwegian and English-Spanish)
and backtranslations into English.
Either the original English sentence or the backtranslation contains a negation, and the other one is an affirmative interpretation.
In this paper, we show that leveraging sentences with negation and their affirmative interpretations
is beneficial for several natural language understanding tasks including natural language inference and sentiment analysis.}
\label{t:motivational-examples}
\end{figure}

In this paper, we address this challenge building upon the following observation:
negation often carries affirmative meanings~\cite{horn1989natural,hasson2006does}. 
For example, people intuitively understand that \emph{John read part of the book} from \emph{John didn't read the whole book}.
Our fundamental idea is to leverage a large collection of sentences containing negation and their affirmative interpretations.
We define an affirmative interpretation as a semantically equivalent sentence that does not contain negation.
We explore this idea by automatically collecting pairs of sentences with negation and their affirmative interpretations
from parallel corpora and backtranslating. Figure~\ref{t:motivational-examples} exemplifies the idea with English-Norwegian and English-Spanish parallel sentences.
Note that
(a)~either the original English sentence or the backtranslation have a negation~(the one that does not is the affirmative interpretation)
and
(b) the meaning of both is equivalent.

Armed with the large collection of sentences containing negation and their affirmative interpretations,
we show that leveraging them yields improvements in three natural language understanding tasks.
First, we address the problem of generating affirmatively interpretations
in the AFIN benchmark~\cite{hossain2022question}, a collection of sentences with negation and their manually curated affirmative interpretations.
Second, we address natural language inference using three common benchmarks:
RTE~\cite{Dagan:2005:PRT:2100045.2100054,
bar2006second,
giampiccolo-etal-2007-third,
bentivogli2009fifth}, 
SNLI~\cite{bowman-etal-2015-large}, and
MNLI~\cite{N18-1101}.
Third, we address sentiment analysis using SST-2~\cite{socher-etal-2013-recursive}.
The main contributions of this paper are:\footnote{Code and data available at \url{https://github.com/mosharafhossain/large-afin-and-nlu}.}
\begin{compactenum}
\item A large collection (153,273) of pairs of sentences containing negation and their affirmative interpretations.
  We present an automated procedure to get these pairs and an analysis of the negation types (single tokens, morphological, lexicalized, etc.).
\item Experimental results with the T5 transformer~\cite{2020t5} showing that blending our pairs during the fine-tuning process
  is beneficial to generate affirmative interpretations from the negations in AFIN.
\item Experimental results showing that a RoBERTa-based classifier~\cite{liu2019roberta} to solve the task of natural language inference benefits from training with new premise-hypothesis derived from our pairs~(two entailments per pair).
\item Experimental results showing that a RoBERTa-based classifier for sentiment analysis benefits from a novel component that automatically generates affirmative interpretations from the input sentence.
\end{compactenum}

The key resource enabling the experimental results is our large collection of pairs of sentences containing negation and their affirmative interpretations.
As we shall see,
the experiments under (2) and (3) are a somewhat straightforward applications of these pairs.
The affirmative interpretation generator we use to improve sentiment analysis, however,
has the potential to improve many natural language understanding tasks.

\section{Related Work}
\label{s:related-work}

Solving natural language understanding tasks when the input text contains negation is challenging.
Researchers have approached negation processing mainly by identifying the scope~\cite{vincze2008bioscope,L12-1077} and 
focus~\cite{blanco-moldovan:2011:ACL-HLT20111}.
Scope refers to the part of the meaning that is negated
and
focus refers to the part of the scope that is most prominently negated~\cite{english.grammar.2002}.
There are many works targeting scope 
detection~\cite{fancellu-etal-2016-neural,fancellu-etal-2017-detecting,li-lu-2018-learning,jumelet-hupkes-2018-language,chen2019attention,zhao-bethard-2020-berts} and 
focus detection~\cite{zou-etal-2014-negation,zou-etal-2015-unsupervised,shen-etal-2019-negative,hossain-etal-2020-predicting}. 
While scope and focus pinpoint what is and what is not negated,
they do not reveal affirmative interpretations as defined in this paper.
Additionally,
there is limited empirical evidence showing that scope or focus is beneficial to solve a natural language understanding task.
\newcite{jimenez} show that scope improves sentiment analysis,
but they do not experiment with modern networks that may not benefit from explicit scope information.

Outside of scope and focus,
\newcite{jiang-etal-2021-im} work with commonsense implications involving negations 
(e.g., ``If X doesn't wear a mask'' then ``X is seen as carefree'').
Closer to our work, 
\newcite{hosseini-etal-2021-understanding} pretrain BERT with an unlikelihood loss
calculated with automatically obtained negated statements.
Their negated statements do not preserve meaning.
The authors show that their method, BERTNOT, outperforms BERT with 
LAMA~\cite{kassner-schutze-2020-negated}
and the same natural language inference corpora we work with.
The work proposed here outperforms theirs (Section \ref{ss:large-afin-nli}) and does not require any manual effort.



We are not the first to work with affirmative interpretations from negated statements.
For example,
\newcite{sarabi-etal-2019-corpus} create a small corpus of verbal negations retrieved from Simple Wikipedia and their affirmative interpretations (total: 5,900).
Simple Wikipedia is a version of Wikipedia that uses shorter sentences and simpler language.
\newcite{hossain2022question} propose a question-answer driven approach to create AFIN, a collection of 3,001 sentences with negation and their affirmative interpretations.
Both of these previous efforts employ humans to collect affirmative interpretations and neither one conducts extrinsic evaluations.
Unlike them,
we automatically collect pairs of sentences with negation and their affirmative interpretations.
Additionally, extrinsic evaluations show that despite our collection procedure is noisy,
leveraging our pairs is beneficial to solve three natural language understanding tasks.


\section{Collecting Sentences with Negation and Their Affirmative Interpretations}
\label{s:large-scale-afin}

This section outlines our approach to create a large collection of sentences containing negation and their affirmative interpretations. 
First, we present the sources of parallel corpora we work with.
Second, we describe our multilingual negation cue detector to identify negation cues in the parallel sentences.
Third, we describe the backtranslation step and a few checks to improve quality.
Lastly, we present an analysis of the resulting sentences with negation and their affirmative interpretations.

\subsection{Selecting Parallel Corpora}
\label{ss:bitexts}

We select parallel sentences in English and either Norwegian or Spanish for two reasons:
(a)~large parallel corpora are available in these language pairs
and 
(b)~negation cue annotations are available in monolingual corpora for the three languages.
The latter is a requirement to build a multilingual cue detector (Section \ref{ss:cue-detection}).
We extract the parallel sentences from three parallel corpora available in the OPUS portal~\cite{tiedemann-2012-parallel}):
WikiMatrix~\cite{schwenk-etal-2021-wikimatrix}, 
CCMatrix~\cite{schwenk-etal-2021-ccmatrix,fan2021beyond}, and 
UNPC~\cite{ziemski-etal-2016-united}. 
Table \ref{t:source-stat} (Column~3) shows the number of parallel sentences we collect from each of the corpora and language pair (total: 14.6 million).

\begin{table}[t]
\small
\centering
\begin{tabular}{l lrrr}
\toprule
              & Source       & \#parl. sents. & \#pairs   & \%pairs\\
\midrule
\multirow{2}{*}{\begin{sideways}en-no\end{sideways}} 
              & WikiMatrix   & 530,000         &  10,274    & 1.94  \\
              & CCMatrix     & 8,000,000       &  73,394    & 0.92  \\
\midrule
\multirow{2}{*}{\begin{sideways}en-es\end{sideways}} 
             & UNPC         & 2,800,000       &  28,028    & 1.00  \\
             & WikiMatrix   & 3,290,000       &  41,577    & 1.26  \\
\midrule 
              & All          & 14,620,000      &  153,273   & 1.05  \\

\bottomrule
\end{tabular}
\caption{
Number of parallel sentences in the English-Norwegian and English-Spanish parallel corpora we work with,
and pairs of sentences with negation and affirmative interpretations we automatically generate via backtranslation.
The yield (\%pairs) is low, but as we shall see
these pairs are useful to solve natural language understanding tasks when negation is present without hurting results when negation is not present.
}
\label{t:source-stat}
\end{table}

\subsection{Identifying Negation Cues in Multiple Languages}
\label{ss:cue-detection}


In order to detect negation in the parallel sentences, 
we develop a multilingual negation cue detector that works with English, Norwegian, and Spanish texts. 
To this end, we fine-tune a multilingual BERT (mBERT)\footnote{\url{https://github.com/google-research/bert/blob/master/multilingual.md}}~\cite{devlin-etal-2019-bert} 
with negation cue annotations in the three languages we work with:
English~\cite{morante-daelemans-2012-conandoyle},  
Norwegian~\cite{maehlum-etal-2021-negation}, and 
Spanish~\cite{jimenez2018sfu}.
We fine-tune jointly for all three languages by combining the original training splits into a multilingual training split.
We terminate the training process after the F1 score in the (combined) development split does not increase for 5 epochs;
the final model is the one which yields the highest F1 score during the training process.
Additional details regarding training procedure and hyperparameters are provided in Appendix \ref{as:cue-detection}.
Our multilingual detector is not perfect but obtains competitive results (F1 scores):
English: 91.96 (test split), Norwegian: 93.40 (test split), and Spanish: 84.41 (dev split, as gold annotations for the test split are not publicly available).
The system detects various negation cue types
including single tokens (no, never, etc.), 
affixal, and lexicalized negations (Section \ref{ss:corpus-analysis}).


We use our multilingual cue detector to detect negation in the 14.6 million of parallel sentences. 
In the English-Norwegian parallel sentences (8.5M), 
negation is present in
both sentences (WikiMatrix: 7.3\%, CCMatrix: 14.2\%),
either sentence (WikiMatrix: 5.2\%, CCMatrix: 5.2\%),
or neither sentence (WikiMatrix: 87.5\%, CCMatrix: 80.6\%).
Similarly, in English-Spanish parallel sentences, 
negation is present in both sentences
(UNPC: 10.7\%, WikiMatrix: 5.7\%),
either sentence (UNPC: 4.6\%, WikiMatrix: 4.4\%),
or neither sentence (UNPC: 84.7\%, WikiMatrix: 89.9\%).
Since we are interested in sentences containing negation and their affirmative interpretations, we only keep the sentences in which either the source or target sentence contains negation.

\begin{table*}[t]
\small
\centering
\begin{tabular}{p{1.65in}l}
\toprule
Negation Type & Examples \\ \midrule

Single tokens (49.6\%)
  & They are still \uline{not} integrated into the German community.\\
\multirow{2}{1.65in}{\emph{Cues: not, n't, no, never, without, nothing, nowhere, nobody, none, etc.}}
  & They have yet to integrate into German society.\\ \addlinespace

& I have \uline{no} doubt that we will reach our goal.  \\ 
& We shall surely get there! \\ \addlinespace

& This process allows for higher precision that could \uline{never} be achieved by hand.  \\ 
& This process allows more precision than anyone performed manually. \\ \midrule

Affixal (30.15\%)
  & The north wing was left largely \uline{un}touched and forms the present house. \\
\emph{Cues: un-, in-, -less, etc.}
  & Only the North wing remained quite intact, and constitutes the current house.  \\ \addlinespace

& We fall in love, and any attempt at logic is use\uline{less}. \\
& We fall in love and any attempt at logic is futile. \\ \midrule

Lexicalized (8.76\%)
  & A further problem was the \uline{lack} of skilled labour.  \\
\emph{Cues: prevent, lack, etc.}
  & Another problem was the issue of obtaining sufficiently qualified personnel. \\ \midrule

Multitoken (2.58\%)
  & After some time, the drainage of water \uline{no longer} occurs. \\
\emph{Cues: no longer, not at all, etc.}
  & After a certain time, the drainage of water ends. \\ \midrule

Multiple negations (8.95\%)
  & The declaration before the courts is \uline{not} valid if the child is \uline{not} 14 years old. \\
  & Any statement in a court is invalid if the child is below 14 years of age. \\ \bottomrule

\end{tabular}

\caption{Examples of sentences with negation and their affirmative interpretations automatically obtained from bitext corpora via backtranslation.
We present examples for several negation types; common \emph{single-tokens} that are not lexicalized negations are the most frequent.
These sentences with negations and their affirmative interpretations come from our collection (Section \ref{s:large-scale-afin}) and include errors.
For example, the affirmative interpretation in the second to last example includes \emph{ends}, a lexicalized negation, because our cue detector did not identify it.}
\label{t:large-afin-examples}
\end{table*}

\subsection{Generating Affirmative Interpretations}
\label{ss:final-pairs}

After identifying negation cues in the parallel sentences,
we backtranslate into English the sentence in the target language (either Norwegian or Spanish; they may or may not contain a negation).
In particular, we utilize Google Cloud Translation API.\footnote{Google Translate API - \url{https://cloud.google.com/translate}}
Before backtranslating, we exclude sentences in the target language if they are longer than 40 tokens,
as longer sentences tend to result in lower translation quality~\cite{fonteyne-etal-2020-literary}.

Backtranslating into English from either Norwegian or Spanish may introduce or remove a negation cue.
We discard such backtranslations
since our goal is to obtain pairs of sentences containing negation and its affirmative interpretation
(i.e., a semantically equivalent sentence that does not contain negation).
The last two columns in Table \ref{t:source-stat} present how many pairs we obtain (total: 153,273).
While the yield is small (1.05\%),
we note that the process is automated and could be expanded to use additional parallel corpora.


\subsection{Quality and Analysis}
\label{ss:data-quality}
The process to collect pairs of sentences with negation and their affirmative interpretations is noisy.
First, the negation cue detector is not perfect thus there are pairs in which the affirmative interpretation contains a negation.
Additionally, backtranslating introduces errors thus the affirmative interpretations are not always semantically equivalent to the sentences containing negation.
Our goal is not to create a gold standard but to collect a large collection 
that we can leverage to improve models for natural language understanding tasks~(Section~\ref{s:experiments-discussion}).

Despite 100\% correctness is not the goal,
we conducted a manual validation with a random sample of 100 pairs.
We discovered that 78\% are correct,
where correct means that the affirmative interpretation satisfies the definition (i.e., no negation and semantically equivalent to the sentence with negation).
We found two main reasons for incorrect pairs.
First, the negation cue detector sometimes fails to detect cues, resulting in affirmative interpretations that contain negation
(e.g.,
\emph{The execution had been \uline{unlawful}}:
\emph{This act would have been \uline{illegal}}; the prefix \emph{il-} is not identified as a negation cue).
Second, the backtranslation sometimes results in the original sentence without the negation cue and thus opposite meanings.
(e.g., English-Norwegian parallel sentences: \emph{How can you \uline{not} enjoy this trip!}, \emph{Hvordan kan du nyte denne turen!}; backtranslation: \emph{How can you enjoy this trip!}).


\paragraph{Analyzing the Negations and Affirmative Interpretations}
\label{ss:corpus-analysis} 
The negation cue detector identifies several types of negation cues.
As a result, our collection of pairs of sentences with negation and their affirmative interpretations includes several negation types (Table \ref{t:large-afin-examples}).
Note that this table presents real examples from our collection including erroneous ones (e.g., some affirmative interpretations contain negation).
The most frequent negation type (49.6\%) are common single-token negation cues such as \emph{not}, \emph{n't} and \emph{never}.
Affixal negations are surprisingly common (30.15\%) and include both prefixes (e.g., \emph{\uline{un}touched}) and suffixes (\emph{use\uline{less}}).
Lexicalized negations usually take the form of a noun (e.g., lack, dismissal) or verb (e.g., prevent, avoid) and account for almost 9\%.
Finally, a few negations (2.58\%) are multitoken (e.g., no longer, not at all),
and several negations (almost 9\%) appear in sentences with at least one more negation.

The corresponding affirmative interpretation is never just the original sentence with negation after removing the negation cue---doing so results in a sentence that is not semantically equivalent.
The required modification are sometimes relatively simple and mainly require swapping a verb or adjective.
For example, 
\emph{are still \uline{not} integrated} becomes \emph{have yet to integrate},
\emph{largely \uline{un}touched} becomes \emph{quite intact}, and
\emph{is use\uline{less}} becomes \emph{is futile}.
Yet the affirmative interpretation often is a more thorough rewrite of the original sentence:
\begin{compactitem}
\item \emph{I have \uline{no} doubt that we will reach our goal.} becomes \emph{We shall surely get there!};
\item \emph{higher precision that could \uline{never} be achieved by hand} becomes \emph{more precision than anyone performed manually; and}
\item \emph{\uline{lack} of skilled labour} becomes \emph{issue of obtaining sufficiently qualified personnel}.
\end{compactitem}

\section{Experiments with Natural Language Understanding Tasks}
\label{s:experiments-discussion}

We leverage our collection of sentences with negation and their affirmative interpretations
to enhance models for three natural language understanding tasks.
First, we leverage them in a blending training setup to
generate affirmative interpretations from the negations in a previous benchmark (Section~\ref{ss:afin-generator}).
Second, we leverage them to create new premise-hypothesis pairs and build more robust models for natural language inference (Section~\ref{ss:large-afin-nli}).
Third, we use them to train a plug-and-play neural component to generate affirmative interpretations from negation.
Then, we incorporate the generator into the task of sentiment analysis (Section \ref{ss:sentiment-analysis}).
We use existing corpora for all tasks as described below;
for natural language inference and sentiment analysis we use the versions released by the GLUE benchmark~\cite{wang-etal-2018-glue}.

Our experimental results show that leveraging the large collection of negations and their affirmative interpretations improves results across all tasks using previously proposed benchmarks.
Specifically, we obtain either slightly better or comparable results when negation is not present in the input,
and always better results when negation is present.


\subsection{Generating Affirmative Interpretations from Negation}
\label{ss:afin-generator}
There are a couple corpora with sentences containing negation and their manually curated affirmative interpretations (Section \ref{s:related-work}).
In our first experiment, we explore whether leveraging our collection is beneficial to generate affirmative interpretations from the negations in AFIN~\cite{hossain2022question}, a manually curated corpus that is publicly available.
AFIN contains 3,001 sentences with negation and their affirmative interpretations.
Unlike our collection (Section \ref{s:large-scale-afin}),
AFIN only considers verbal negations (i.e., the negation cues always modify a verb).
Here are some examples:
\begin{compactitem}
\item \emph{It was \uline{not} formed by a natural process.}\\
  \emph{It was formed by an artificial process.}
\item \emph{An extinct volcano is one that has \uline{not} erupted in recent
history.}\\
  \emph{An extinct volcano erupted in the past.}
\end{compactitem}

The AFIN authors experiment with the T5~\cite{2020t5} transformer to automatically generate affirmative interpretations.
While the task remains a challenge and our results are much worse than the human upper bound,
we show that incorporating our collection of sentences containing negation and their affirmative interpretations during the training process
results in a more robust generator.


\paragraph{Blending Our Collection of Negations and Affirmative Interpretations}
\label{p:blended-training}
We adopt a blending technique by \newcite{shnarch-etal-2018-will}
in order to maximize the chances that the training process benefits from our collection of negations and affirmative interpretations. 
Since our collection is much larger than the training split in AFIN  (153k vs. 2.1k),
simply adding our collection to the training split and fine-tuning T5 as usual would result in a model that underperforms with AFIN.
There are three phases in the training process.
In the first phase, we fine-tune T5 with the combination of our collection and the training split in AFIN for $m$ epochs.
In the second phase, we continue to fine-tune T5 blending our collection and the training split in AFIN for $n$ epochs.
  The blending factor ($[0..1]$) determines the number of instances from our collection that we incorporate in each epoch.
  This number decreases after each epoch.
In the third phase,
we fine-tune using the training split in AFIN for $k$ epochs.
We refer the reader to Appendix \ref{as:affirmative-intp-generation} for additional details on the training process and hyperparameters.

\begin{table}
\small
\centering
\begin{tabular}{l cccc}
\toprule
& BLEU & chrf++ & METEOR \\ \midrule 
T5 transformer     & 26.5 & 50.5 & 43.5 \\
~~~+ blending Ours & 28.6 & 52.5 & 45.8 \\
\bottomrule
\end{tabular}
\caption{Automatic evaluation of the T5 transformer on the task of generating affirmative interpretations as defined in the AFIN benchmark~\cite{hossain2022question}.
Blending our pairs in the process of fine-tuning T5 yields improvements with all metrics.
}
\label{t:automatic-metrics}
\end{table}

\begin{table}
\small
\centering
\setlength{\tabcolsep}{0.072in}

\begin{tabular}{l r r r r r}
\toprule
     & \multicolumn{5}{c}{Validation Scores} \\ \cmidrule{2-6}
     &    \multicolumn{1}{c}{4}    &   \multicolumn{1}{c}{3}     &   \multicolumn{1}{c}{2}     &   \multicolumn{1}{c}{1}   &   \multicolumn{1}{c}{0}   \\ 
\midrule
Upper Bound        &   86.2  &   11.6  &    2.0   &   0.2   &   n/a  \\ \midrule
T5 transformer     &   32.0  &   15.3  &   12.0   &   3.3   &  37.3  \\
~~~+ blending Ours &   37.0  &   14.0  &   13.0   &   10.0   &  26.0  \\
\bottomrule

\end{tabular}
\caption{
Manual evaluation of the T5 transformer on the task of generating affirmative interpretations as defined in the AFIN benchmark.
The upper bound comes from the manual validation by the creators of AFIN.
Blending our pairs in the fine-tuning process generates more correct interpretations (higher validation scores are better).
}
\label{t:human-eval-score}
\end{table}

\paragraph{Results and Discussion}
\label{p:afin-results-and-discussion}
Table \ref{t:automatic-metrics} presents the evaluation with the test split in AFIN using automatic metrics:
BLEU-2~\cite{papineni2002bleu}, 
chrf++~\cite{popovic2017chrf++}, and
METEOR~\cite{banerjee2005meteor}.
We obtained these scores comparing the gold affirmative interpretations in AFIN and the predicted ones by T5.
Despite our collection of negations and affirmative interpretation is noisy, out-of-domain, and considers more negation types,
leveraging it is beneficial.
Indeed, we observe improvements across the three metrics (BLEU-2: 28.6 vs. 26.5, chrf++: 52.5 vs. 50.5, and METEOR: 45.8 vs. 43.5). 

\begin{table*}
\small
\centering
\setlength{\tabcolsep}{.07in}
\begin{tabular}{lc ccc ccc ccc}
\toprule
 &&
 \multicolumn{2}{c}{RTE}  &&
 \multicolumn{2}{c}{SNLI} &&
 \multicolumn{2}{c}{MNLI} \\
 \cmidrule(lr){3-4} \cmidrule(lr){6-7} \cmidrule(lr){9-10}
              && dev & neg. P-H && dev & neg. P-H && dev  & neg. P-H \\ \midrule
              
w/o negation fine-tuning \\
~~~BERT~\cite{devlin-etal-2019-bert}    && 66.10 & 57.60 && 89.90 & 44.40 && 83.20 & 63.90 \\
~~~XLNet~\cite{yang2019xlnet}   && 69.90 & 60.90 && 90.60 & 51.50 && 86.70 & 66.30 \\
~~~RoBERTa~\cite{liu2019roberta} && 75.80 & 62.50 && 91.60 & 51.90 && 87.90 & 66.70 \\ \midrule

w/ negation fine-tuning \\
~~~BERTNOT~\cite{hosseini-etal-2021-understanding} && 69.68 & 74.47 && 89.00 & 45.96 && 84.31 & 60.89 \\
~~~RoBERTa blending Ours                           && 77.62 & 78.13 && 91.35 & 54.87 && 87.00 & 67.89 \\

\bottomrule
\end{tabular}

\caption{
Results (accuracy) using several transformers and
(a)~the development splits in RTE, SNLI, and MNLI,
and
(b)~ the \emph{new} premise-hypothesis containing negation (neg. P-H) from \citet{hossain-etal-2020-analysis}.
RoBERTa blending new premise-hypothesis derived from our sentences with negation and their affirmative interpretations substantially outperforms
the three transformers without any negation fine-tuning and BERTNOT with the new premise-hypothesis that contain negation while obtaining comparable results with the original development splits.
}
\label{t:results-nli}
\end{table*}

\paragraph{Manual Validation}
\label{p:afin-qualitative-analysis}
Automatic metrics for generation tasks are useful but have well-known limitations~\cite{mathur-etal-2020-tangled,bert-score}.
Following the AFIN authors,
we also conduct a manual evaluation.
Specifically, we validate a sample of 100 automatically generated affirmative interpretations.
The validation consists in assigning a score indicating how confident they are in the correctness of an affirmative interpretation given the sentence containing negation~(4: extremely confident, 
3: very confident, 
2: moderately confident, 
1: slightly confident).
We also include 0 to indicate that the affirmative interpretations is wrong.
We show examples of each score in Appendix \ref{as:affirmative-intp-generation}. 

Table \ref{t:human-eval-score} shows the manual evaluation.
Blending our collection of negations and affirmative interpretations yields better results.
While still far from the upper bound,
blending increases the confidence scores. 
Most notably, the percentage of incorrect affirmative interpretations decreases from 37.3\% to 26.0\% ($\Delta =-30.3\%$).

\subsection{Natural Language Inference}
\label{ss:large-afin-nli}
Our collection of pairs of sentences with negation and their affirmative interpretations can be seen as
semantically equivalent sentences in which only one statement contains negation.
By definition, there are two entailment relationships between semantically equivalent sentences---using either sentence as premise and the other one as hypothesis. 
We thus create two premise-hypothesis sets from each pair in our collection and label them as \emph{entailment} to create a large collection of entailments involving negation.
For example, we generate the following entailments from the pair
(\emph{The universal nature of these rights and freedoms does \uline{not} admit doubts},
\emph{The universal nature of these rights and freedoms is beyond question}):
\begin{compactitem}
\item Premise: \emph{The universal nature of these rights and freedoms does \uline{not} admit doubts.}\\
   Hypothesis: \emph{The universal nature of these rights and freedoms is beyond question.}
\item Premise: \emph{The universal nature of these rights and freedoms is beyond question.}\\
   Hypothesis: \emph{The universal nature of these rights and freedoms does \uline{not} admit doubts.}
\end{compactitem}

This process results in 306,546 new premise-hypothesis annotated \emph{entailment} (2 per pair in our collection)
without any manual effort.

We experiment with
(a)~three transformer-based classifiers without any fine-tuning designed to improve results when there is a negation in the premise or hypothesis,
(b)~BERTNOT~\cite{hosseini-etal-2021-understanding}, a BERT transformer pretrained with a modified loss calculated in part with automatically obtained negated statements,
and
(c)~a RoBERTa-based classifier blending our 306k new \emph{entailment} premise-hypothesis using the strategy presented in Section \ref{ss:afin-generator}.
We refer the reader to Appendix \ref{as:nli} for additional details about the models, training process, and hyperparameters.
Regarding corpora, we work with 
RTE~\cite{Dagan:2005:PRT:2100045.2100054,
bar2006second,
giampiccolo-etal-2007-third,
bentivogli2009fifth}, 
SNLI~\cite{bowman-etal-2015-large}, and
MNLI~\cite{N18-1101}.
Additionally,
we work with the 4,500 new premise-hypothesis pairs by \newcite{hossain-etal-2020-analysis},
who derive them from RTE, SNLI and MNLI by adding a negation to a premise, hypothesis or both.

\paragraph{Results and Discussion}
Table \ref{t:results-nli} presents the results.
We train all models with the corresponding training split, except \emph{RoBERTa blending Ours}, which also blends our 306k new premise-hypothesis pairs during the training process.
We present results with the corresponding development split (gold labels for the test split are not available for all them) and the premise-hypothesis including negation.
We find that blending our 306k premise-hypothesis is beneficial despite these pairs
(a)~only include entailments
and
(b)~inherit the errors present in our collection of sentences with negation and their affirmative interpretations.
With the original development splits in RTE, SNLI, and MNLI,
we either obtain slightly better results (RTE,~+1.82) or comparable (SNLI:~-0.25, MNLI: -0.90).
The improvements are consistent, however, with the pairs that include negation (neg. P-H).
RTE and SNLI benefit the most (78.13 vs. 62.50, 54.87 vs. 51.90).
We hypothesize that MNLI benefits the least (67.89 vs. 66.70)
because premises in MNLI are often multiple sentences and our new premise-hypothesis are always single sentences.

\begin{figure}[t]
\centering
\includegraphics[width=0.99\columnwidth]{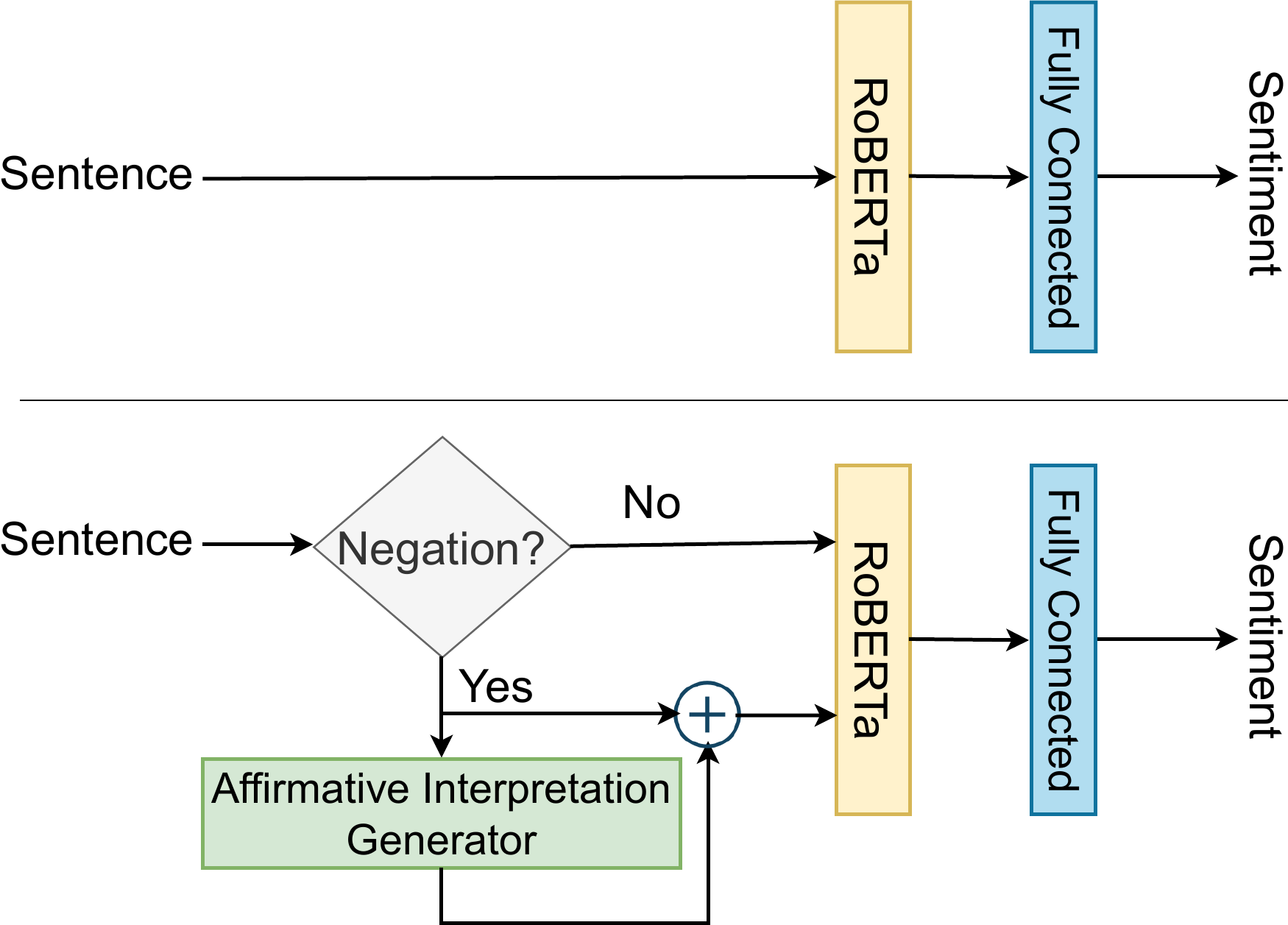} 
\caption{
Standard architecture for a transformer-based classifier (top)
and
modification to include our affirmative interpretation generator (bottom).
If a sentence contains a negation, we provide RoBERTa with the original sentence and the affirmative interpretation.
The generator is pretrained with our large collection of sentences with negation and their affirmative interpretations.
}
\label{f:sentiment-networks}
\end{figure}

\subsection{Sentiment Analysis}
\label{ss:sentiment-analysis}
We close our experiments exploring the task of sentiment analysis
(i.e., classifying sentences according to their sentiment: positive or negative).
Our motivation is that state-of-the-art systems for sentiment analysis face challenges with negation~\cite{ribeiro-etal-2020-beyond}.
Unlike generating affirmative interpretations~(Section~\ref{ss:afin-generator}) and natural language inference~(Section~\ref{ss:large-afin-nli}),
however, it is unclear how to leverage our large collection of sentences with negation and their affirmative interpretations to alleviate the issue.

We propose a task-agnostic solution: complement input sentences containing negation with their automatically generated affirmative interpretations.
Figure \ref{f:sentiment-networks} illustrates this proposal in the architecture at the bottom.
Rather than feeding all sentences to a transformer-based classifier (top),
we first check whether input sentences contain a negation with our cue detector (Section \ref{s:large-scale-afin}).
If they do not, we feed them to the classifier as usual.
If they do, we
(a)~automatically generate its affirmative interpretation with the generator described in Section \ref{ss:afin-generator}
and
(b)~feed to the transformer-based classifier both the original sentence with negation and the affirmative interpretation separated by the [SEP] token. 
Appendix \ref{as:sentiment-analysis} provides additional details about the training process. 

We experiment here with RoBERTa as it produces very competitive results.
Note that our strategy to complement negated inputs with their affirmative interpretations
could be used with any classifier for any task as long as it takes a text as its input.
Regarding corpora,
we use 
\label{p:sentiment-experiment-setup}
SST-2~\cite{socher-etal-2013-recursive} as released by GLUE~\cite{wang-etal-2018-glue}.
It consists of 70,042 movie reviews and sentiment annotations for each sentence.

Here are a few examples of automatically generated affirmative interpretations from sentences containing negation in SST-2:
\begin{compactitem}
\item \emph{It is \uline{not} a bad film.} \\
   Affirmative interpretation: \emph{It is a good movie.}
\item \emph{She may \uline{not} be real, but the laughs are.}\\
   Affirmative interpretation: \emph{She is fictional.}
\item \emph{He feels like a spectator and \uline{not} a participant}\\
   Affirmative interpretation: \emph{He feels like a spectator rather than participant.}
\item \emph{The movie has \uline{no} idea of it is serious.}\\
   Affirmative interpretation: \emph{The movie has a lack of idea that it is serious.}
\item \emph{A thriller \uline{without} a lot of thrills.}\\
   Aff. interpretation: \emph{A thriller with little thrills.}
\end{compactitem}

Note that they are by no means perfect, but they mostly preserve meaning while not using negation. 
For example, the second affirmative interpretation only covers the meaning of part of the original sentence with negation (i.e., \emph{She may not be real}),
and the second to last includes a negation (\emph{lack}).


\begin{table}
\small
\centering
\newcolumntype{P}[1]{>{\centering\arraybackslash}p{#1}}
\begin{tabular}{lP{.40in}P{1in}}
\toprule
                    & \multicolumn{2}{c}{affirmative intpn. generator?} \\ \cmidrule{2-3}
                    & No & Yes \\ \midrule

dev. w/o neg.       & 94.0        & 94.7 (+0.7\%) \\
dev. w/  neg.       & 93.0        & 94.8 (+1.9\%) \\
\midrule
~~~important negs.   & 86.0        & 89.8 (+4.4\%) \\
~~~unimportant negs. & 95.0        & 95.8 (+0.8\%) \\
\bottomrule

\end{tabular}

\caption{Results (macro F1) with RoBERTa using the SST-2 development split.
We provide results with instances that contain and do not contain negation as well as important and unimportant negations.
Our affirmative interpretation generator yields improvements across the board,
especially with instances containing important negations
(i.e., when removing the negation changes the sentiment polarity (positive or negative).
}
\label{t:results-sentiment}
\end{table}

\paragraph{Results and Discussion}
\label{p:sentiment-results-discussion}

Table \ref{t:results-sentiment} presents the results (macro F1).
We provide results with the sentences in the development split depending on whether they contain a negation (gold labels for the test split are not publicly available).
Additionally, we use the grouping of negations by~\citet{hossain-etal-2022-analysis}: important or unimportant.
A negation is unimportant if removing it does not change the sentiment (e.g., both
\emph{I got a headache watching this meaning\uline{less} downer}
and
\emph{I got a headache watching this downer} are negative.

Incorporating our affirmative interpretation generator is always beneficial.
This includes instances containing negation (94.8 vs. 93.0) and, surprisingly,
instances that do not contain negation (94.7 vs. 94.0).
We hypothesize that this is the case because our negation cue detector is not perfect thus instances are sometimes fed through the incorrect branch after the \emph{Negation?} fork (Figure \ref{f:sentiment-networks}).
As one would expect, the generator makes the classifier more robust with important negations (89.8 vs. 86),
but we also observe improvements with unimportant negations (95.8 vs. 95.0).


\section{Conclusions}
\label{s:conclusion}

Negation poses a challenge for natural language understanding.
Understanding negation requires humans to infer affirmative meanings~\cite{horn1989natural}
(e.g., \emph{The lot has not been vacant} conveys \emph{The lot has been occupied}).
Inspired by this insight,
we collect a large collection (153k) of pairs of
sentences containing negation and their affirmative interpretations.
We define the latter as a semantically equivalent sentence that does not contain negation.
Our collection process relies on parallel corpora and backtranslation and is automated.

We show that leveraging our collection is beneficial to solve three natural language understanding tasks:
(a)~generating affirmative interpretations,
(b)~natural language inference,
and
(c)~sentiment analysis.
All our experiments use out-of-domain, manually curated corpora.
Crucially, our proposal yields better results when negation is present in the input while slightly improving or obtaining comparable results when it is not.
Additionally, our proposal does not require any manual annotations.

\section*{Limitations}
\label{s:limitations}

In order to create a large collection of sentences with negation and their affirmative interpretations, 
we use publicly available parallel sentences.
We note that in two of the three sources (i.e., WikiMatrix~\cite{schwenk-etal-2021-wikimatrix} and CCMatrix~\cite{schwenk-etal-2021-ccmatrix,fan2021beyond}), the authors use auto-alignment methodologies to collect the parallel sentences. This step may introduce errors in the original sources. 
Next, to detect negation cues in the huge collections of parallel sentences (14.6 millions), we develop a multilingual cue detection system that is certainly not 100\% perfect. 
While the cue detector performs well on the negation corpora it is trained with (Section \ref{ss:cue-detection}), some incorrect predictions can be expected in the parallel corpora we use.
Furthermore, the translation API introduces additional noise backtranslating from Norwegian or Spanish into English (Section \ref{ss:final-pairs}). 
Regarding models and experiments, we leverage RoBERTa and T5 (Section \ref{s:experiments-discussion}) 
as systems based on them perform well on natural language understanding tasks.\footnote{\url{https://super.gluebenchmark.com/leaderboard}}
However, we acknowledge that other transformers such as XLNet \cite{yang2019xlnet} and DeBERTa \cite{he2021deberta} may yield better results.

\section*{Acknowledgements}
This material is based upon work supported by the National Science Foundation under 
Grant No.~1845757.
Any opinions, findings, and conclusions or recommendations expressed in this material
are those of the authors and do not necessarily reflect the views of the NSF.
Computational resources were provided by the UNT office of High-Performance Computing. 
Further, we leveraged computational resources from the Chameleon platform \cite{keahey2020lessons}. 
We also thank the reviewers for insightful comments.

\bibliography{refs}
\bibliographystyle{acl_natbib}

\appendix

\section{Additional Details on Identifying Negation Cues in Multiple Languages}
\label{as:cue-detection} 
Referring to Section \ref{ss:cue-detection} of the paper, we employ an off-the-shelf multilingual BERT-Base model (cased version) pretrained on 104 languages.\footnote{\url{https://github.com/google-research/bert/blob/master/multilingual.md}}
We concatenate the contextualized representations from the last and third-to-last layers.
Then, we pass the concatenation to a fully connected layer. 
Finally, we leverage a conditional random field (CRF) layer that yields the output sequence identifying negation cues.
Since a negation cue can consist of multiple tokens (e.g., \emph{by no means}), we use the BIO (B: Beginning, I: Inside, and O: Outside) tagging scheme. 
The system takes 1.5 hours on average to train on a single core NVIDIA Tesla V100 (32GB). 
Table \ref{t:cue-detector-params} lists the tuned hyperparameters for the cue detector. 
We avail the code for all our experiments at \url{https://github.com/mosharafhossain/large-afin-and-nlu}.

\begin{table}[t]
\small
\centering
\begin{tabular}{lc}
\toprule
Hyperparameter           \\
\midrule
Maximum Epochs    &  25 \\
Batch Size        &  10 \\
Patience          &  5   \\
Maximum sentence length   & 150 \\
Optimizer         & AdamW \\
Learning rate (mBERT)     & 1e-5 \\
Learning rate (FC and CRF)     & 1e-3 \\
Weight decay (mBERT)      & 1e-5  \\
Weight decay (FC)      & 1e-3  \\
Dropout (mBERT)        & 0.5 \\
Gradient clipping    & 5.0 \\
Warmup epochs      & 5 \\
Accumulate step   & 1 \\
\bottomrule

\end{tabular}
\caption{
Hyperparameters for finetuning the multilingual cue detector (Section \ref{ss:cue-detection} in the paper). FC refers to Fully Connected layer.
}
\label{t:cue-detector-params}
\end{table}
\begin{table}[t!]
\small
\centering
\begin{tabular}{lc}
\toprule
Hyperparameter           \\
\midrule
Maximum Epochs        &  20 \\
Batch Size        &  8 \\
Patience          &  5   \\
Input sentence length (max.)   & 80 \\
Target sentence length (max.)  & 50 \\
Optimizer         & Adafactor \\
Learning rate     & 1e-5 \\
Weight decay      & 1e-6  \\
Gradient clipping     & 5.0 \\
Warmup epochs      & 3 \\
Accumulate step   & 1 \\
top\_k             & 50 \\
top\_p             & 0.95 \\
repetition\_penalty   & 2.5 \\
\bottomrule

\end{tabular}
\caption{
Hyperparameters for finetuning our affirmative interpretation generator (Section \ref{ss:afin-generator} in the paper).
}
\label{t:appendix-affirm-interp}
\end{table}

\begin{table*}[t!]
\small
\centering
\begin{tabular}{p{1.2in}p{4.65in}}
\toprule
 & Examples \\ \midrule

\multirow{2}{1.2in}{Extremely confident \\(Score: 4)}
  & Move them to low places so that they do \uline{not} fall.\\
  & Affirmative Interpretation: They fall when they are in higher places.\\ \addlinespace

  & The ones that were \uline{not} rewarded were \uline{not} marked with fields.  \\ 
  & Affirmative Interpretation: The ones that were rewarded were marked with fields. \\ \midrule

\multirow{2}{1.2in}{Very confident \\(Score: 3)}
  & The most recent successful bids for the Olympic and Paralympic Games were in cities that had \uline{never} hosted them before. \\
  & Affirmative Interpretation: Other cities had hosted them once.\\ \addlinespace

  & \uline{No} other studies could find a link between the vaccine and autism. \\ 
  & Affirmative Interpretation: A study found a link between the vaccine and autism. \\ \midrule
  
\multirow{2}{1.2in}{Moderately confident \\(Score: 2)}
  & In 1984, because the games were in New York, and because of the boycott, from when we boycotted in 1980, \uline{not} a lot of European countries came over. \\
  & Affirmative Interpretation: Lots of European countries came over in 1980. \\ \addlinespace

  & It occurs when the body does \uline{not} receive enough iron. \\ 
  & Affirmative Interpretation: The body receives too little iron.\\  \midrule
  
\multirow{2}{1.2in}{Slightly confident \\(Score: 1)}
  & I understand third party candidates have \uline{no} success.\\
  & Affirmative Interpretation: Third party candidates have minimal success. \\ \addlinespace

  & I do\uline{n't} expect that the lack of British participation will stop any action. \\ 
  & Affirmative Interpretation: I expect that the lack of British participation will slow down any action.\\  \midrule
  
\multirow{2}{1.2in}{Wrong affirmative interpretation \\(Score: 0)}
  & I have throughout my career \uline{not} supported needle exchanges as anti-drug policies.\\
  & I have supported needle exchanges as anti-drug policies. \\ \addlinespace

  & Unlike other organelles, the ribosome is \uline{not} surrounded by a membrane. \\ 
  & The ribosome is surrounded by a membrane.\\  
 
\bottomrule

\end{tabular}

\caption{
Examples of sentences containing negation from AFIN and their affirmative interpretations automatically generated. The generator uses T5 trained with our large collection of sentences with negation and their affirmative interpretations (Section \ref{ss:afin-generator}). We show examples of the manual validation; scores range from 0 to 4.
}
\label{t:t5-generated-examples}
\end{table*}

\section{Additional Details on Generating Affirmative Interpretations from Negation}
\label{as:affirmative-intp-generation}

We utilize the Huggingface implementation \cite{wolf-etal-2020-transformers} of T5, a conditional generation model (Section \ref{ss:afin-generator}). 
In each run, the system requires approximately 7.2 hours to train on a single core NVIDIA Tesla V100 (32GB). 
Table \ref{t:appendix-affirm-interp} shows the hyperparameters for this experiment. 
Regarding  Section \ref{p:afin-qualitative-analysis} in the paper (Manual Validation), we present examples of each score in Table \ref{t:t5-generated-examples}.

\begin{table*}[t]
\small
\centering
\begin{tabular}{l r r r}
\toprule
      Hyperparameter    &    RTE       &   SNLI    & MNLI \\ 
\midrule

Maximum epochs           &    10        &   8       &  8 \\ 
Warmup epochs            &    4         &   2       &  3 \\ 
Batch size               &    24        &   16      &  10 \\ 
Patience                 &    5         &   5       &  5 \\ 
Maximum sentence length            &    100       &   80      &  100 \\ 
Optimizer                & AdamW        &   AdamW   &  AdamW \\
Learning rate            & 1e-5         &   1e-5    &  1e-5\\
Weight decay             & 0.0          &   5e-6    &  5e-6\\
Gradient clipping           & 5.0          &   5.0     & 5.0\\
Dropout                  & 0.2          &   0.3     & 0.3\\

\bottomrule
\end{tabular}
\caption{
Hyperparameters for finetuning RoBERTa with blending our pairs and the NLI benchmarks (Section \ref{ss:large-afin-nli} in the paper).
}
\label{t:appendix-nli}
\end{table*}

\section{Additional Details on Solving Natural Language Inference}
\label{as:nli}

We evaluate the systems on the development splits of the NLI benchmarks we work with (Section \ref{ss:large-afin-nli}) as test split labels are not publicly available.  
So, we randomly select 15\% examples of the original training split in order to tune the hyperparameters and to select the best model during the training process for each benchmark.  
We note that we evaluate on the development split with matched genres for MNLI. 
In each run, the system (blending with ours) requires approximately 2.1 hours to train for RTE, 7.8 hours for SNLI, and 9.6 hours for MNLI on a single core NVIDIA Tesla V100 (32GB).
Table \ref{t:appendix-nli} presents the hyperparameters we use in this experiment. 

\begin{table}[t]
\small
\centering
\begin{tabular}{lc}
\toprule
Hyperparameter           \\
\midrule
Maximum Epochs        &  5 \\
Batch Size        &  16 \\
Patience          &  3   \\
Maximum sentence length   & 80 \\
Optimizer         & AdamW \\
Learning rate     & 1e-5 \\
Weight decay      & 0.0  \\
Dropout           & 0.5 \\
Gradient clipping     & 5.0 \\
Warmup epochs      & 5 \\
Accumulate step   & 1 \\
\bottomrule

\end{tabular}
\caption{
Hyperparameters for finetuning our SST-2 system presented in Section \ref{ss:sentiment-analysis} in the paper.
}
\label{t:appendix-sentiment-analysis}
\end{table}

\section{Additional Details on Solving Sentiment Analysis}
\label{as:sentiment-analysis}

To experiment with SST-2, we randomly select 5\% examples of the original training split for tuning the hyperparameters as well as for selecting the best model during the training process since test labels are not publicly available in SST-2 (part of GLUE \cite{wang-etal-2018-glue}). 
On average,  the system takes half an hour to train on a single core NVIDIA Tesla V100 (32GB). 
We share the tuned hyperparameters in Table \ref{t:appendix-sentiment-analysis}.

\end{document}